# COVID-19 Infodemic – Understanding content features in detecting fake news using a machine learning approach

**Vimala Balakrishnan[1], Hii Lee Zing[2], Eric Laporte[3]**
[1,2]Faculty of Computer Science & Information Technology, Universiti Malaya, 50603, Kuala Lumpur, Malaysia
[3]LIGM, Univ Gustave Eiffel, CNRS, F-77454 Marne-la-Vallée, France
Email: vimala.balakrishnan@um.edu.my[1](corresponding author), leezing0303@gmail.com[2], eric.laporte@univ-paris-est.fr[3]

**Abstract**

The use of content features, particularly textual and linguistic for fake news detection is under-researched, despite empirical evidence showing the features could contribute to differentiating real and fake news. To this end, this study investigates a selection of content features such as word bigrams, part of speech distribution etc. to improve fake news detection. We performed a series of experiments on a new dataset gathered during the COVID-19 pandemic and using Decision Tree, K-Nearest Neighbor, Logistic Regression, Support Vector Machine and Random Forest. Random Forest yielded the best results, followed closely by Support Vector Machine, across all setups. In general, both the textual and linguistic features were found to improve fake news detection when used separately, however, combining them into a single model did not improve the detection significantly. Differences were also noted between the use of bigrams and part of speech tags. The study shows that textual and linguistic features can be used successfully in detecting fake news using the traditional machine learning approach as opposed to deep learning.

## 1. Introduction

Over the past two decades, the growth of Internet and emergence of social media and text messaging platforms such as Facebook, Twitter, WhatsApp etc. has brought dramatic changes to the way people communicate with each other, easing all restrictions (Faustini & Covões, 2020). However, this also brought on another issue – the wide dissemination of unverified news, and especially of fake news. Fake news can be defined as "news articles that are intentionally and verifiably false and could mislead leaders" (Allcott & Gentzkow, 2017), but in fact people posting and sharing fake news are not necessarily aware that the content is false. Fake news often encompasses misinformation, disinformation and malinformation (Woetzel, 2021; Zhang & Ghorbani, 2020). Types of fake news vary as well, for example, clickbait is a type of news that exaggerates its headlines to attract more users to read the articles, often with no match between the content and the headlines (Woetzel, 2021) whereas satire mimics real news for humorous intent with more slang, curse words and clauses to enhance the comedic effect (Bondielli & Marcelloni, 2019; Rubin et al., 2016).

Although not a new concept or phenomenon, the dissemination of fake news has increased dramatically during the COVID-19 period (Elhadad et al., 2020; van der Linden et al., 2020; Torpan et al., 2021). Reports on the spread of COVID-19 related fake news are many, ranging from those about the virus itself, to those about home-made remedies, vaccination and its side-effects, and local Standard Operating Procedures (SOP), among others. The problem became so worrisome that several measures were applied to mitigate the spread: for example, the World Health Organization (WHO) requested popular search engines like Google, Yahoo and other platforms to display their official reports and information related to COVID-19 as top hits (Shu & Shieber, 2020). Improving digital literacy among the public is a part of the solution, but the spread of fake news can also be mitigated using Artificial Intelligence (AI), particularly machine learning techniques. Fact-checking websites such as PoliFact.com, Snopes.com, and Sebenarnya.my (a Malaysian fact-checking portal) are commonly proposed to address this issue, however, this solution is considered laborious as users and/or authorities are required to manually check the authenticity of dubious news or claims. Although this approach has the advantage of focusing on news articles individually, it tends to be costly and it would be impractical to implement it on a large scale, considering the volume of news that is published and shared every day (Ahmed et al., 2018; Huang & Chen, 2020).

### 1.1 Fake news detection

A comprehensive literature review revealed that fake news detection studies have mostly attempted to improve the detection using machine learning algorithms such as Random Forest, Support Vector Machine (SVM), and Logistic Regression etc. with promising results (Chowdhury et al., 2020; Kudugunta & Ferrara, 2018; Purnomo et al., 2017; Sicilia et al., 2018). For example, Sicilia et al. (2018) developed a Zika-related rumor detection model using Random Forest with 71% accuracy while Purnomo et al. (2017) classified fake news using Random Forest in Snopes.com with an accuracy of 95.95%. The majority of the studies focused on English-language corpora (Ahmed et al., 2018; Huang & Chen, 2020; Purnomo et al., 2017; Sommariva et al., 2018), and very few targeted other languages such as Chinese (Liu et al., 2019), Portuguese (Monteiro et al., 2018) and Bulgarian (Hardalov et al., 2016).

Scholars made further attempts to improve fake news detection by including features as auxiliary input to the algorithm in addition to the news content. For instance, Shu et al. (2019) explored characteristics of users who are more likely to share fake news, using network-related features (number of followers, number of posts, number of favorites etc.) and objective demographics likely to be correlated with fake news sharing (e.g., age, personality, location). The authors reported 90.9% and 96.6% accuracy using Random Forest on the PolitiFact and GossipCop datasets, respectively. On the other hand, Chowdhury et al. (2020) computed a jointly learned credibility score of both publishers and users who shared news, using Logistic Regression, with results indicating an accuracy of 91.3% and 85.8% on Politifact and Buzzfeed datasets, respectively. A more recent study investigated the use of Facebook users' profile features (e.g., profile picture, number of page likes, news posts etc.) and news content (e.g., headline, body text, images etc.) to detect fake news using traditional machine learning as well as deep learning algorithms (Sahoo & Gupta, 2021). The authors obtained the best results (i.e., 99.4% accuracy) with Long Short-Term Memory (LSTM) and both types of features combined. A similar study based on features extracted from the headlines and news content, using the Extreme Gradient Boosting (XGBoost) algorithm optimized by the Whale Optimization algorithm, reported an accuracy of 91.8% (Sheikhi, 2021).

Other recent studies have shown that online deceptions can be detected based on language and writing formats of the news content features, particularly textual and linguistic features, i.e. features extracted from the text of the news (Bondielli & Marcelloni, 2019; Parikh & Atrey, 2018). Commonly used textual and linguistic features include character-level features such as proportion of uppercase letters or punctuation symbols, word-level features (type/token ratio—TTR, number of characters per word, frequency of unique words, word bigrams etc.), sentence-level features such as sentence length or number of sentences, and linguistic features belonging to shallow syntax (parts of speech—POS, e.g. noun, adjective, verb), deep syntax (grammatical structure of the sentences) or semantics (emotional charge, polarity, semantic classes of words) (Conroy et al., 2015; Tompkins, 2019). For instance, empirical evidence exists that fake news tends to be less complicated, often containing shorter sentences (Mahyoob et al., 2020). However, studies specifically using these features to improve automatic detection of fake news are scarce. For example, Abonizio and colleagues (2020) extracted features related to complexity (average words per sentence, TTR, word size etc.), style (POS, uppercase letters, quotation marks etc.) and psychology (i.e., sentiment) to identify fake news in English, Portuguese and Spanish, using several machine learning algorithms including K-Nearest Neighbors (KNN), SVM, Random Forest and XGBoost. The authors found the combination of the complexity- and style-related features produced the best results. In another recent study by Faustini and Covões (2020), the authors computed various textual and linguistic features such as proportion of uppercase characters, of exclamation marks, of adverbs, of nouns, number of sentences, and number of characters, Word2Vec representation and spelling errors, among others, on five different datasets. Binary fake news detection (i.e., real versus fake) models were developed using Naïve Bayes, KNN, SVM and Random Forest with results showing the proportion of exclamation and question marks were of little help while the number of unique words and polarity were very helpful in fake news detection. Both SVM and Random Forest emerged as the best algorithms.

Others such as Bharadwaj and Shao (2019) used Term Frequency Inverse Document Frequency (TF-IDF) and *n*-grams as textual features to train fake news detection models using Naïve Bayes and Random Forest, with results indicating that bigrams outperform unigrams, trigrams, and 4-grams, with an accuracy score of 90.77% and 95.66% for Naïve Bayes and Random Forest, respectively. Rubin and colleagues (2016) used SVM along with five features, namely, absurdity, humor, grammar, negative affect, and punctuation to detect fake news among 360 news articles, with findings indicating the best combination of these features can detect satirical news with an 87% F-score. Monteiro et al. (2018) used POS tags, semantic classes, Bag of Words (BoW), proportion of punctuation symbols, emotiveness, uncertainty, and non-immediacy (defined with reference to 1st, 2nd and 3rd person) as features to detect fake news. The authors found BoW to achieve an accuracy of 88%, followed by POS tags (75%) and semantic classes (73%). Table 1 provides a summary of studies that have used textual and linguistic features with machine learning algorithms, both involving corpora in English and in other languages.

Table 1. Summary of works on fake news detection using machine learning with textual and linguistic features

| Author, Purpose | Dataset | Feature Extraction Technique | Features | Algorithms | Best results |
|---|---|---|---|---|---|
| (Elhadad et al., 2020): To detect misleading information related to COVID-19 | Information sources from WHO, UNICEF and UN | TF, TF-IDF (Uni-, Bi-, Tri-, N-Gram), Word Embeddings | N-gram | DT, MNB, BNB, LR, KNN, Perceptron, NN, SVM and ERF, XGBoost | NN - 99.68% accuracy. |
| (Abonizio et al., 2020): To develop a multi-class model to classify news into fake, genuine and satire | News written in English, Portuguese and Spanish | Statistical analysis, POS, Named Entity Recognition, Sentiment | average words per sentence, word size; count of sentences; TTR; POS-tag diversity; ratio of Named Entities, quotation marks, ADJ-, ADP-, ADV-, DET-, NOUN-, PRON-, PROPN-, PUNCT-, SYM-, VERB- tag; Uppercase letters; OOV words frequency; sentiment polarity | K-NN, SVM, RF and XGBoost | RF - 85.3% accuracy.<br><br>Complexity + Stylometric had the best performance. |
| (Faustini & Covões, 2020): To detect fake news from different platforms and languages | Text and social media posts in three languages: Germanic, Latin and Slavic | Statistical analysis, Word2Vec, BoW, Document-Class Distance | proportion of uppercase characters, exclamation marks, NOUN, ADV, ADJ, question marks; exclamation marks; number of unique words, sentences, characters; words per sentence, sentiment of message, Word2Vec n | KNN, NB, RF, SVM | RF - 95% accuracy.<br><br>SVM – 94% accuracy<br><br>BoW achieved the best result. |
| (Monteiro et al., 2018): To detect fake news written in Portuguese | News written in Portuguese | POS, BoW | POS semantic classes; BoW; pausality; emotiveness; uncertainty; non-immediacy | SVM, NB, RF, MLPN | SVM - 89% accuracy.<br><br>BoW and emotiveness achieved the best result. |
| (Bharadwaj & Shao, 2019): To detect fake news written in Portuguese | Kaggle | TF, TF-IDF (Uni-, Bi-, Tri-, Quad-gram) | semantic features: N-grams | NB, RF | RF - 95.66% accuracy.<br><br>Bigram produced the best results |

Note: ADJ: Adjectives; ADP: Adposition; ADV: Adverbs; BNB: Bernoulli Naïve Bayes; BoW: Bag-of-Words; DET: Determiner; DT: Decision Tree; ERF: Ensemble Random Forest; K-NN: K-Nearest Neighbor; LR: Logistic Regression; MLPN: Multilayer Perceptron; MNB: Multinomial Naïve Bayes; NB: Naïve Bayes; NN: Neural Network; OOV: Out-of-Vocabulary; POS: Part-of-Speech; PRON: Pronoun; PROPN; Personal Pronouns; PUNCT = Punctuation; RF: Random Forest; SVM: Support Vector Machine; SYM: Synonyms; TF: Term-Frequency; TF-IDF: Term Frequency-Inverse Document Frequency; TTR: Type-Token Ratio UN: United Nations; UNICEF: United Nations Children's Fund; WHO: World Health Organization; XGB: Extreme Gradient Boosting

The literature review led us to the following research question: How to build an effective fake news detection model incorporating textual and linguistic features using machine learning? The present study answers this question by extracting such features to train fake news detection models using machine learning algorithms identified from the literature. The study evaluates the proposed models on a new fake news dataset gathered from Malaysia's fact-checking website, namely, sebenarnya.my, targeting COVID-19 related (fake) news.

## 2. Materials and Method

The overall architecture used in the present study is depicted in Figure 1, showing five main phases, namely, data collection, pre-processing, feature extraction, fake news detection and model evaluation. Phases 2 and 3 occur in parallel, not sequentially.

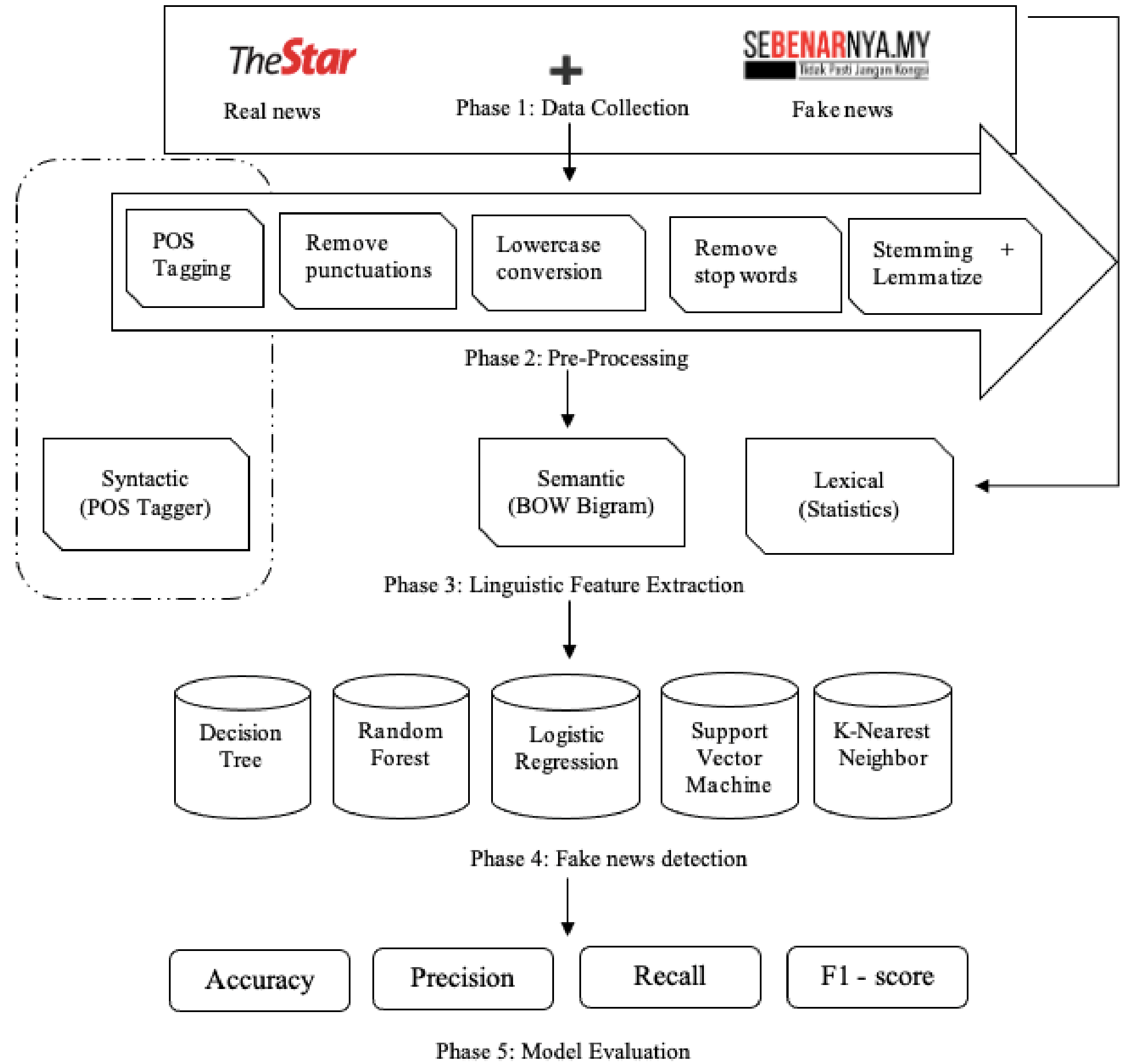

Figure 1: Overall fake news detection process

### 2.1 Data collection

The fake news dataset was developed based on COVID-19 fake news identified in Malaysia between January 2020 and March 2021 (sebenarnya.my portal). The fake news articles in the dataset contain textual content in English and standard Malay, date of posting, location, URL of where the news was published, and theme classifications such as health, crime, general etc. It has been checked that the content is fake by the relevant authorities. As this source for the dataset contains only fake news, it has been necessary to include news articles with another label, for a binary classification mechanism to work. Therefore, following the steps adopted in previous studies (Faustini & Covões, 2020; Rubin et al., 2016; Silva et al., 2020) in developing fake news corpora, a semi-automatic process was used to crawl a local news website (i.e., The Star: COVID-19 Watch) for true counterparts of the fake news, using specific keywords. Specifically, this second source of news was filtered using "Covid-19" and "Nation = Malaysia", with the timeline ranging between January 2020 and March 2021. In total, there were 1,512 news articles comprising 699 fake and 813 real news articles (see Figure 2 for samples of fake and real news). Thus, this resulted in a balanced dataset; no class imbalance issue exists. The dataset size is limited by the fact that the fake news is extracted from the fact-checking portal. This process allowed us to work on a verified dataset, hence the falsity of the fake news is established, without resorting to costly human annotation. Previous fake news detection studies have worked on smaller (e.g., (Rubin et al., 2016) with 360 news articles) or similar numbers of samples (e.g., (Abonizio et al., 2020) with 421 fake news articles in Spanish and 846 in Portuguese).

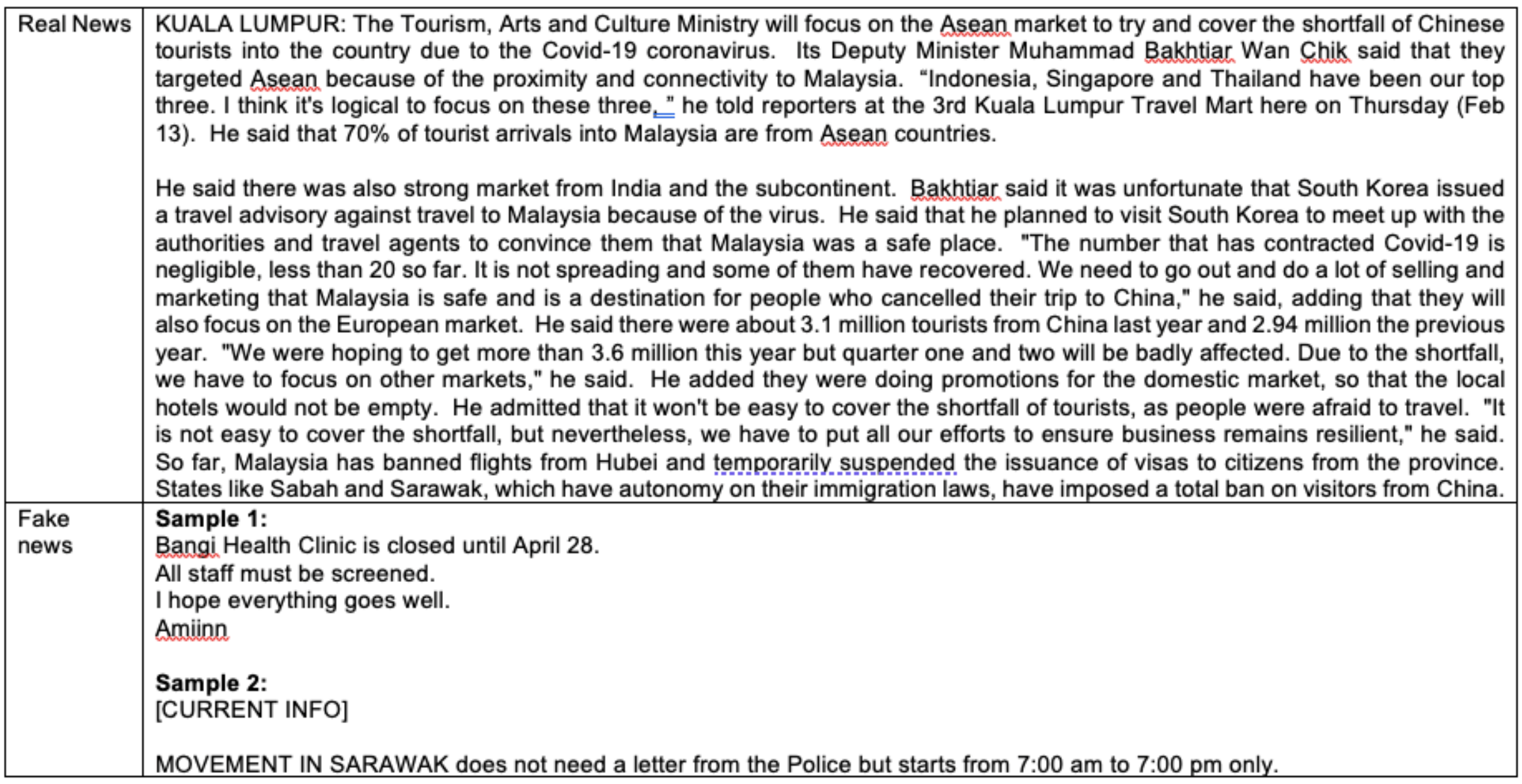

| | |
|---|---|
| Real News | KUALA LUMPUR: The Tourism, Arts and Culture Ministry will focus on the Asean market to try and cover the shortfall of Chinese tourists into the country due to the Covid-19 coronavirus. Its Deputy Minister Muhammad Bakhtiar Wan Chik said that they targeted Asean because of the proximity and connectivity to Malaysia. "Indonesia, Singapore and Thailand have been our top three. I think it's logical to focus on these three," he told reporters at the 3rd Kuala Lumpur Travel Mart here on Thursday (Feb 13). He said that 70% of tourist arrivals into Malaysia are from Asean countries.<br><br>He said there was also strong market from India and the subcontinent. Bakhtiar said it was unfortunate that South Korea issued a travel advisory against travel to Malaysia because of the virus. He said that he planned to visit South Korea to meet up with the authorities and travel agents to convince them that Malaysia was a safe place. "The number that has contracted Covid-19 is negligible, less than 20 so far. It is not spreading and some of them have recovered. We need to go out and do a lot of selling and marketing that Malaysia is safe and is a destination for people who cancelled their trip to China," he said, adding that they will also focus on the European market. He said there were about 3.1 million tourists from China last year and 2.94 million the previous year. "We were hoping to get more than 3.6 million this year but quarter one and two will be badly affected. Due to the shortfall, we have to focus on other markets," he said. He added they were doing promotions for the domestic market, so that the local hotels would not be empty. He admitted that it won't be easy to cover the shortfall of tourists, as people were afraid to travel. "It is not easy to cover the shortfall, but nevertheless, we have to put all our efforts to ensure business remains resilient," he said. So far, Malaysia has banned flights from Hubei and temporarily suspended the issuance of visas to citizens from the province. States like Sabah and Sarawak, which have autonomy on their immigration laws, have imposed a total ban on visitors from China. |
| Fake news | **Sample 1:**<br>Bangi Health Clinic is closed until April 28.<br>All staff must be screened.<br>I hope everything goes well.<br>Amiinn<br><br>**Sample 2:**<br>[CURRENT INFO]<br><br>MOVEMENT IN SARAWAK does not need a letter from the Police but starts from 7:00 am to 7:00 pm only. |

Figure 2: Sample fake and real news

## 2.2 Data pre-processing

Several pre-processing steps were deployed prior to detecting fake news, including translating all texts written in standard Malay into English, to be able to compare our results with previously published work on English corpora. In addition, Malay language processing faces challenges due to lack of text corpora, POS taggers and other resources (Lan, Logeswaran, 2020). The texts were translated by Google Translator and verified by a linguistic expert, who corrected and improved some of the translations. In cases where the fake news was in the form of images, only the corresponding text was extracted (i.e., no images or emojis). Other standard Natural Language Processing (NLP) tasks followed this process, including POS tagging, removal of emoticons, special characters, and stop words, conversion of uppercase letters to lowercase, stemming and lemmatization (i.e., identifying root words) using the Standard Core NLP parser.

2.3 Textual and linguistic feature extraction

We selected nine textual and linguistic features (see Table 2 for the explanation of each extracted feature) and divided them in three sets by singling out (i) the POS statistics, which belong to shallow syntax, and (ii) the bigrams, the most informative feature in our selection. The other seven features are considered to be correlated with complexity and style and make up our third set of features. POS tagging was run first to avoid breaking the sequence of the words during the pre-processing steps (Elhadad et al., 2020). The final feature extracted was the semantic feature (i.e., bigrams), accomplished using BoW.

Table 2: Details of each feature

| Textual and linguistic features | Features | Explanation |
|---|---|---|
| Character-level (Mosallam et al., 2014) | Average number of uppercase letters | The number of uppercase letters divided by the number of characters in the news |
| | Average number of punctuation symbols | The number of punctuation characters divided by the number of characters in the news |
| | Average number of numeric characters | The number of numeric digits divided by the number of characters in the news. |
| Word-level | Average number of stop words | The number of stop words divided by the number of words in the news |
| | | |
| | | |
| | Type-Token Ratio | The number of unique words divided by the total number of word tokens in the news. The closer the TTR ratio is to 1, the greater the lexical richness of the segment. |
| | Bigrams | Pairs of two consecutive words from a given sentence, e.g., “luxury people”, “forgive us”, etc. |
| Sentence-level | Average length per news | The number of sentences in the news |
| | Average words per sentence | The number of words divided by the number of sentences in the news |
| Syntactic | Ratio of Part of Speech tags | The number of occurrences of each POS tag divided by the total number of occurrences of POS tags |

2.4 Fake news detection models

The proposed fake news detection models were trained using five machine learning algorithms, namely, Decision Tree, Random Forest, Logistic Regression, KNN and SVM. Decision Tree is a learning algorithm used to solve classification and regression problems and known to produce high accuracy for simple datasets (Garg & Goyal, 2020). However, it is prone to the overfitting problem, whereby small changes made in the training data would cause large changes to the decision logic. Conversely, Random Forest is one of the most efficient and reliable ensemble classifiers: it combines several decision trees and selects a final prediction, usually based on a majority voting over the trees (Garg & Goyal, 2020).

Logistic Regression is a classification algorithm based on estimated probability (Bharti et al., 2020). The algorithm assumes that the relevant variables are independent (i.e., no correlation between the variables) and uses a sigmoid function to map any real input values to values between 0 and 1. Another well-known classification algorithm is SVM, which works by finding a hyperplane in an $n$-dimensional space by maximizing the distance between data points (Reddy et al., 2019). The dimensionality $n$ of the space depends on the number of features in the model. The data points that are closest to the hyperplane determine the position and orientation of the hyperplane and are called support vectors. Once the hyperplane is defined, it becomes the decision boundary that classifies data points. The class will be predicted based on the side of the hyperplane where the data point has fallen. One of the advantages of SVM is that it is effective in cases where the number of dimensions is greater than the number of samples (Sathyanarayana & Amarappa, 2014).

Finally, KNN is also based on the distance between data points in a vector space. It assumes that similar things or equivalent label classes are distributed close to one another (Garg & Goyal, 2020). Many methods can be used to calculate the distance, but the most popular and familiar method is the Euclidean distance. Selecting the correct *k* parameter is important for tuning the algorithm to a task. Odd numbers are often used to have a tie breaker when there are two classes, and normally a higher *k* value will have a more accurate prediction up to a certain extent (Garg & Goyal, 2020). All these machine learning algorithms have been used in previous fake news studies (as shown in Table 1), with Random Forest and SVM consistently emerging as the best classifier.

2.5 Evaluation and experiments

The linguistic-based fake news detection model was evaluated in various setups as described below:

- $Baseline_X$: refers to the execution of the classification algorithm based on text only, i.e., without any features (X refers to the algorithm used).
- $Bigram_X$: refers to the Baseline model above, however with the inclusion of bigrams
- $POS_X$: refers to the Baseline model above, however with the inclusion of POS statistics
- $Misc_X$: refers to the Baseline model above, however with the inclusion of the 7 style- and complexity-related features
- $All_X$ – Baseline model with the inclusion of all textual and linguistic features

All the models above were trained and tested using a 10-fold cross validation, an unbiased approach as every observation from the dataset has a chance to become a part of train and test sets (Raschka, 2018). The approach is also deemed appropriate considering the small dataset. All the models were evaluated using the standard metrics for text classification: accuracy (i.e., proportion of correctly classified fake news items among all news items), recall (proportion of actual fake news items identified as such), precision (proportion of correctly classified news items among those classified as fake news) and F1-score (harmonic mean between recall and precision). All the metrics are reported in %, with higher measures indicating better performance (Alonso et al., 2021).

All the data pre-processing, modelling and evaluation was performed using Python, an open-source scripting language, and the scikit-learn toolkit. Most of the hyperparameter settings were kept with default values, except for *k* for KNN (i.e., $k = 10$) and linear kernel for SVM.

## Results and Discussion

### 2.6 Performance of fake news detection models based on textual and linguistic features

Table 3: Results of the linguistic-based fake news detection models

| Algorithms | Model | 10-Fold cross-validation | | | |
|---|---|---|---|---|---|
| | | Accuracy | Recall | Precision | F1-score |
| Decision Tree | $\text{Baseline}_{DT}$ | 0.9286 | 0.9192 | 0.9062 | 0.9111 |
| | $\text{POS}_{DT}$ | 0.9352 | 0.9227 | 0.9188 | 0.9186 |
| | $\text{Bigram}_{DT}$ | 0.9431 | 0.9363 | 0.9244 | 0.9285 |
| | $\text{Misc}_{DT}$ | 0.9285 | 0.9093 | 0.9133 | 0.9101 |
| | $\text{All}_{DT}$ | 0.9339 | 0.9272 | 0.9145 | 0.9185 |
| Random Forest | $\text{Baseline}_{RF}$ | 0.9569 | 0.9264 | 0.9843 | 0.9537 |
| | $\text{POS}_{RF}$ | 0.9669 | 0.9290 | 0.9911 | 0.9586 |
| | $\text{Bigram}_{RF}$ | 0.9629 | 0.9262 | 0.9942 | 0.9558 |
| | $\text{Misc}_{RF}$ | 0.9656 | 0.9261 | 0.9910 | 0.9572 |
| | $\text{All}_{RF}$ | 0.9669 | 0.9305 | 0.9910 | 0.9593 |
| Logistic Regression | $\text{Baseline}_{LR}$ | 0.9683 | 0.9324 | 0.9910 | 0.9605 |
| | $\text{POS}_{LR}$ | 0.9616 | 0.9180 | 0.9900 | 0.9515 |
| | $\text{Bigram}_{LR}$ | 0.8956 | 0.9646 | 0.8146 | 0.8810 |
| | $\text{Misc}_{LR}$ | 0.9339 | 0.8798 | 0.9582 | 0.9158 |
| | $\text{All}_{LR}$ | 0.9180 | 0.8945 | 0.9049 | 0.8974 |
| SVM | $\text{Baseline}_{SVM}$ | 0.9735 | 0.9557 | 0.9793 | 0.9668 |
| | $\text{POS}_{SVM}$ | 0.9748 | 0.9493 | 0.9896 | 0.9683 |
| | $\text{Bigram}_{SVM}$ | 0.8956 | 0.9682 | 0.8130 | 0.8813 |
| | $\text{Misc}_{SVM}$ | 0.9576 | 0.9225 | 0.9774 | 0.9484 |
| | $\text{All}_{SVM}$ | 0.9325 | 0.9401 | 0.9001 | 0.9175 |
| KNN | $\text{Baseline}_{KNN}$ | 0.8874 | 0.7466 | 0.9688 | 0.8413 |
| | $\text{POS}_{KNN}$ | 0.8756 | 0.7019 | 0.9917 | 0.8206 |
| | $\text{Bigram}_{KNN}$ | 0.4074 | 0.6070 | 0.4058 | 0.5755 |
| | $\text{Misc}_{KNN}$ | 0.9194 | 0.8636 | 0.9365 | 0.8972 |
| | $\text{All}_{KNN}$ | 0.7289 | 0.9556 | 0.6068 | 0.7392 |

Table 3 depicts the accuracy, recall, precision, and F1-score for all the models based on the machine learning algorithms used. Most of the machine learning algorithms performed well in detecting fake news, scoring more than 90% accuracy, except for KNN. In particular, the $\text{Bigram}_{KNN}$ model performed very poorly (i.e., accuracy – 40%; F1-score – 58%), indicating the algorithm is not able to process this feature effectively for detection purposes. The low precision score (i.e., 40.58%) indicates less than half of the fake news articles in the dataset were detected correctly, hence our findings do not recommend the combination of KNN and bigrams.

Random Forest outperformed the other algorithms, followed very closely by SVM for all the features (except for POS), with both accuracy and F1-score more than 95%, which is consistent with most of the studies in Table 1, which found the same two algorithms to yield the best accuracy, regardless of the languages and features involved (Abonizio et al., 2020; Bharadwaj & Shao, 2019; Faustini & Covões, 2020; Monteiro et al., 2018). If we compare our results with previous ones, Random Forest and SVM yielded the best accuracy when syntactic features were used (i.e., 96.7% and 97.5%, respectively), much higher than those reported in Abonizio et al. (2020) using Random Forest (85.3%) and Monteiro et al. (2018) using SVM (88%). These two previous studies, however, are not readily comparable with ours, since they were performed on Spanish and Portuguese, not on English.

The models with the Misc features were also found to perform well, especially Random Forest (accuracy – 96.6% and F1-score – 95.8%) and SVM (accuracy - 95.8% and F1-score - 94.8%). These figures were also higher than those reported in the literature (Abonizio et al., 2020; Faustini & Covões, 2020). For instance, Abonizio and colleagues (2020) examined the effect of a similar selection of textual features (i.e., number of words, sentences, TTR, uppercase, punctuation, quotation marks etc.) with results showing their model performs best (85.3%) when these features were incorporated together. Our findings along with studies on Spanish and Portuguese indicate that textual features improve fake news detection.

The set of bigrams was found to improve fake news detection for Random Forest (accuracy – 96.3%; F1-score – 95.4%) and Decision Tree (accuracy – 94.3%; F1-score – 92.9%). However, a similar effect was not observed for the other algorithms, which were outperformed by their respective baseline models when the bigram feature was included. Bharadwaj and Shao (2019) revealed improved performance when bigrams were used, reporting an accuracy of 95.7% (slightly lower than ours) using Random Forest; however, the authors used the frequency of bigrams, as opposed to the set of bigrams in this study. The decline in the fake news detection performance using SVM and Logistic Regression (compared to the corresponding baseline) could be attributed to the heavy informative load of the set of bigrams, which may require a large training corpus. This, therefore, warrants further investigation.

Interestingly, combining all three sets of textual and linguistic features did not significantly improve the performance of the individual models, maybe indicating that each set affect the detection differently. A similar pattern was reflected in Abonizio et al. (2020) who found their model combining textual and psychological features did not improve fake news detection significantly compared to textual features. In fact, our model with all features performed worse than the baseline for Logistic Regression and SVM, the same two algorithms that obtained lower scores with the bigram feature than with the baseline. This suggests that the inclusion of the bigram feature in the $All_{SVM}$ and $All_{LR}$ models may have contributed to the decline in the detection performance.

## 2.7 Observations on the features

Through informal observation, we noticed several correlations between the values of the textual and linguistic features and the genuineness vs. fakeness of the news.

*3.2.1 The POS distribution feature*

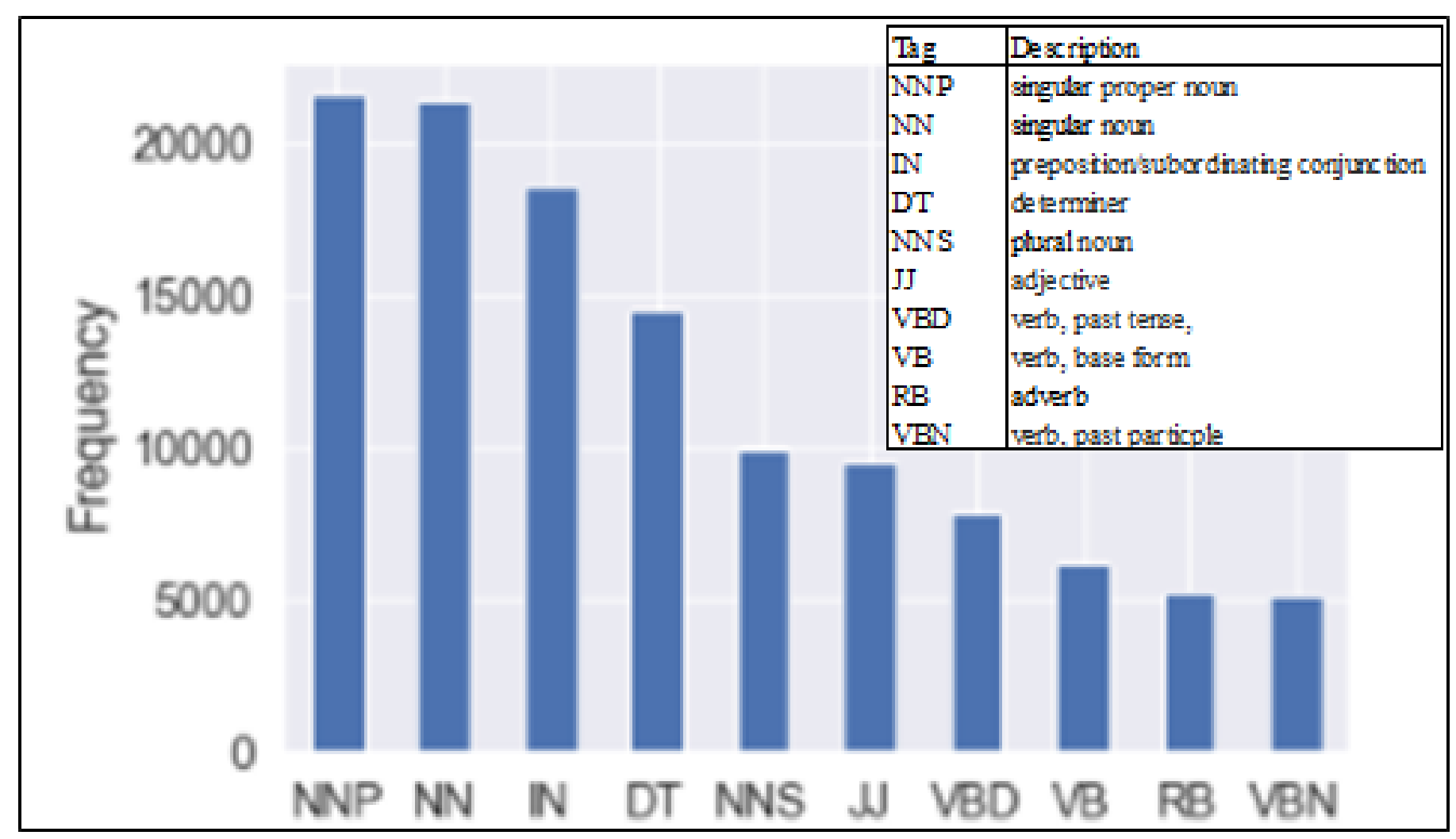

Figure 3a. POS for real news

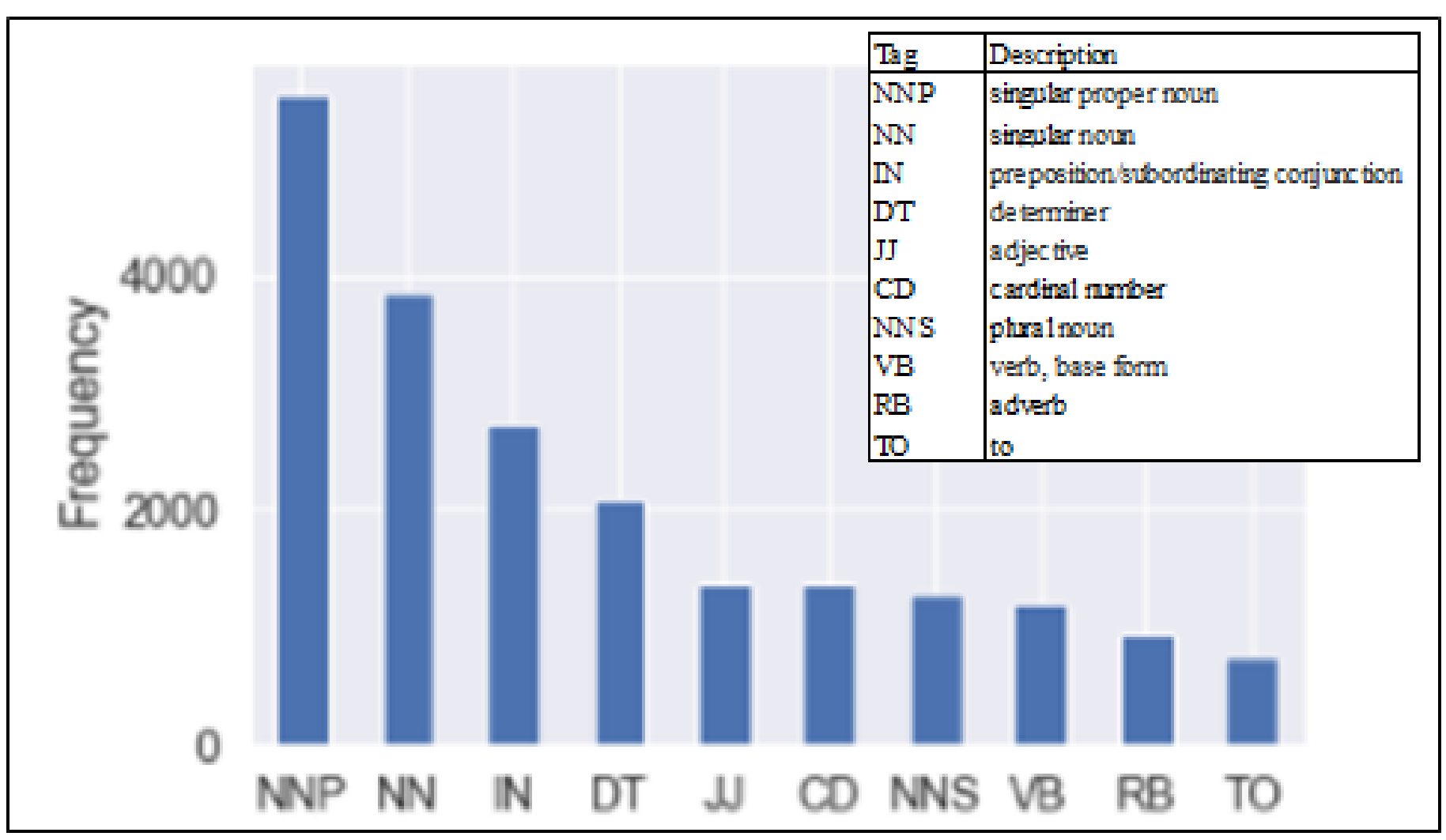

Figure 3b. POS for fake news

Figure 3. POS distributions for real and fake news in the dataset

Figure 3 illustrates the POS breakdown for the real and fake news. It can be observed that the POS tag distributions were similar for both types of news, whereby the Singular Proper Noun (NNP), Singular Noun (NN), Preposition (IN) and Determiner (DT) ranked as the most frequently used four POS tags. However, the Cardinal Digit (CD) and the Infinite Marker (TO) were used more frequently in fake news compared to real news, the latter of which contained more occurrences of past tenses in the forms of verbs in the past tense and in the past participle.

*3.2.2 Bigrams*

The bigrams were extracted using the BoWs technique. Table 4 shows the top 20 bigrams found in the real and fake news.

Table 4: Top 20 bigram for real and fake news

| Number | Real news | Fake news | Number | Real news | Fake news |
|---|---|---|---|---|---|
| 1 | datuk seri | march 2020 | 11 | tan sri | integrated rb |
| 2 | noor hisham | covid 19 | 12 | order mco | kuala lumpur |
| 3 | dr noor | control order | 13 | covid19 pandemic | hari raya |
| 4 | kuala lumpur | movement control | 14 | minister datuk | stay home |
| 5 | petaling jaya | peace upon | 15 | ismail sabri | red zone |
| 6 | movement control | face mask | 16 | tested positive | police station |
| 7 | control order | family member | 17 | face mask | migrant worker |
| 8 | health ministry | corona virus | 18 | standard operating | indonesian migrant |
| 9 | covid19 case | wear mask | 19 | operating procedure | health clinic |
| 10 | prime minister | shah alam | 20 | said statement | positive covid19 |

In real news, the Director-General of Health wad mentioned frequently. *Datuk Seri*, *noor hisham* and *dr noor* (items 1-3) were all referring to the same person, Datuk Seri Dr Noor Hisham, who is the current Director-General of Health in the country. Bigrams such as *control order*, *movement control*, *covid19* appeared in the top 20 bigrams for both the real and fake news, a pattern that was somewhat expected since the dataset under study contains news about the pandemic. However, most of the top 10 bigrams for the real news denote administrative concepts such as the Director-General of Health and the Ministry, as opposed to those for the fake news, several of which are related to practical consequences of the pandemic and lockdown.

*3.2.3 Other features*

An uppercase initial is generally used in the first word of a sentence and in a proper name (Anwar Siddiqui, 2018). The use of uppercase in the real news dataset was found to be higher than in the fake news, indicating a more traditional writing style and/or a higher proportion of proper names in the real news. In fact, this observation tallies with the top 20 bigrams found in real news (Section 3.2.2), several of which denote official entities. Further, it was also found that real news articles contain more sentences and are generally longer. This is in accordance with previous empirical findings showing fake news tend to be less complicated, comprising shorter words and sentences (Horne & Adali, 2017; Mahyoob et al., 2020). However, our finding is in contrast to Kapusta et al. (2020) who found fake news to contain longer sentences. Our analysis also revealed a higher occurrence of stop words in the real news than in the fake news, consistent with others such as Horne and Adali (2017). Further, fake news in the dataset was also found to contain a higher TTR than real news, indicating that the former has a higher degree of lexical variation, echoing results in Abonizio et al. (2020).

## 3. Conclusion, limitations, and future directions

This study explored the use of three selections of textual and linguistic features, specifically (i) bigrams, (ii) POS distribution, and (iii) miscellaneous features related to complexity and style, in order to improve fake news detection using a dataset containing news about the on-going COVID-19 pandemic. Machine learning algorithms were used to develop the models, with results indicating Random Forest and SVM are the best models in detecting fake news with sets of features (i) and (iii) separately, whereas Random Forest and Decision Tree are the best models with set (ii) separately, and also with all three sets together. However, combining all three linguistic features into a single model did not show any significant improvement.

We identify several limitations. First, the source of our fake news data was a local fact-checking portal containing news mostly from social media such as Facebook and WhatsApp etc., whereas the real news was sourced from articles published on the local news website. Therefore, it is inevitable that real news articles were longer, on average, than fake news articles, and written in a more formal style. Future studies could source fake news data in more traditional style to test the specific ability of the models to detect fake news. Second, the dataset used was relatively small, hence it is recommended to replicate the study with a larger dataset, especially for the models that involve the set of bigrams, which contains richer information than the other features. Third, future studies could also investigate original news instead of translated news, as translation is usually not perfect and produces text in a variety of English with unknown properties. Although all the necessary measures were taken to ensure the quality of the translation, it is possible for errors to have occurred, hence affecting the bigram and POS features, for example. Fourth, future studies could explore other linguistic features including sentiment polarity, emotional charge, readability, uncertainty etc. Finally, we used a binary classification (i.e., fake versus real). It would be interesting to see if the linguistic features could be used to differentiate types of fake news (e.g., clickbait, satire, propaganda) as well.

## 4. Funding

This research did not receive any specific grant from funding agencies in the public, commercial, or not-for-profit sectors.